\documentclass[sigconf]{acmart}
\AtBeginDocument{%
  \providecommand\BibTeX{{%
    \normalfont B\kern-0.5em{\scshape i\kern-0.25em b}\kern-0.8em\TeX}}}

\copyrightyear{2022} 
\acmYear{2022} 
\setcopyright{acmcopyright}
\acmConference[SIGIR '22]{Proceedings of the 45th International ACM SIGIR Conference on Research and Development in Information Retrieval}{July 11--15, 2022}{Madrid, Spain.}
\acmBooktitle{Proceedings of the 45th International ACM SIGIR Conference on Research and Development in Information Retrieval (SIGIR '22), July 11--15, 2022, Madrid, Spain}
\acmPrice{15.00}
\acmISBN{978-1-4503-8732-3/22/07} 
\acmDOI{10.1145/3477495.3532047}

\usepackage{multirow}
\usepackage{enumitem}

\renewcommand\footnotetextcopyrightpermission[1]{} 
\begin{document}
\fancyhead{}
\title{Progressive Learning for Image Retrieval \\ with Hybrid-Modality Queries}

\author{Yida Zhao\footnotemark[1], Yuqing Song\footnotemark[1], Qin Jin\footnotemark[2]}
\affiliation{%
 \institution{School of Information, Renmin University of China}
 \country{}
}
\email{{zyiday,syuqing,qjin}@ruc.edu.cn}

\renewcommand{\shortauthors}{Yida Zhao, Yuqing Song, and Qin Jin}
\renewcommand{\authors}{Yida Zhao, Yuqing Song, and Qin Jin}




\begin{abstract}
Image retrieval with hybrid-modality queries, also known as composing text and image for image retrieval (CTI-IR), is a retrieval task where the search intention is expressed in a more complex query format, involving both vision and text modalities. For example, a target product image is searched using a reference product image along with text about changing certain attributes of the reference image as the query. It is a more challenging image retrieval task that requires both semantic space learning and cross-modal fusion.
Previous approaches that attempt to deal with both aspects achieve unsatisfactory performance.
In this paper, we decompose the CTI-IR task into a three-stage learning problem to progressively learn the complex knowledge for image retrieval with hybrid-modality queries.
We first leverage the semantic embedding space for open-domain image-text retrieval, and then transfer the learned knowledge to the fashion-domain with fashion-related pre-training tasks.
Finally, we enhance the pre-trained model from single-query to hybrid-modality query for the CTI-IR task.
Furthermore, as the contribution of individual modality in the hybrid-modality query varies for different retrieval scenarios, we propose a self-supervised adaptive weighting strategy to dynamically determine the importance of image and text in the hybrid-modality query for better retrieval.
Extensive experiments show that our proposed model significantly outperforms state-of-the-art methods in the mean of Recall@K by 24.9\% and 9.5\% on the Fashion-IQ and Shoes benchmark datasets respectively.

\footnotetext[1]{Equal Contribution.}
\footnotetext[2]{Corresponding Author.}
\end{abstract}



\maketitle

\section{Introduction}
\begin{figure}
  \centering
  \includegraphics[width=\linewidth]{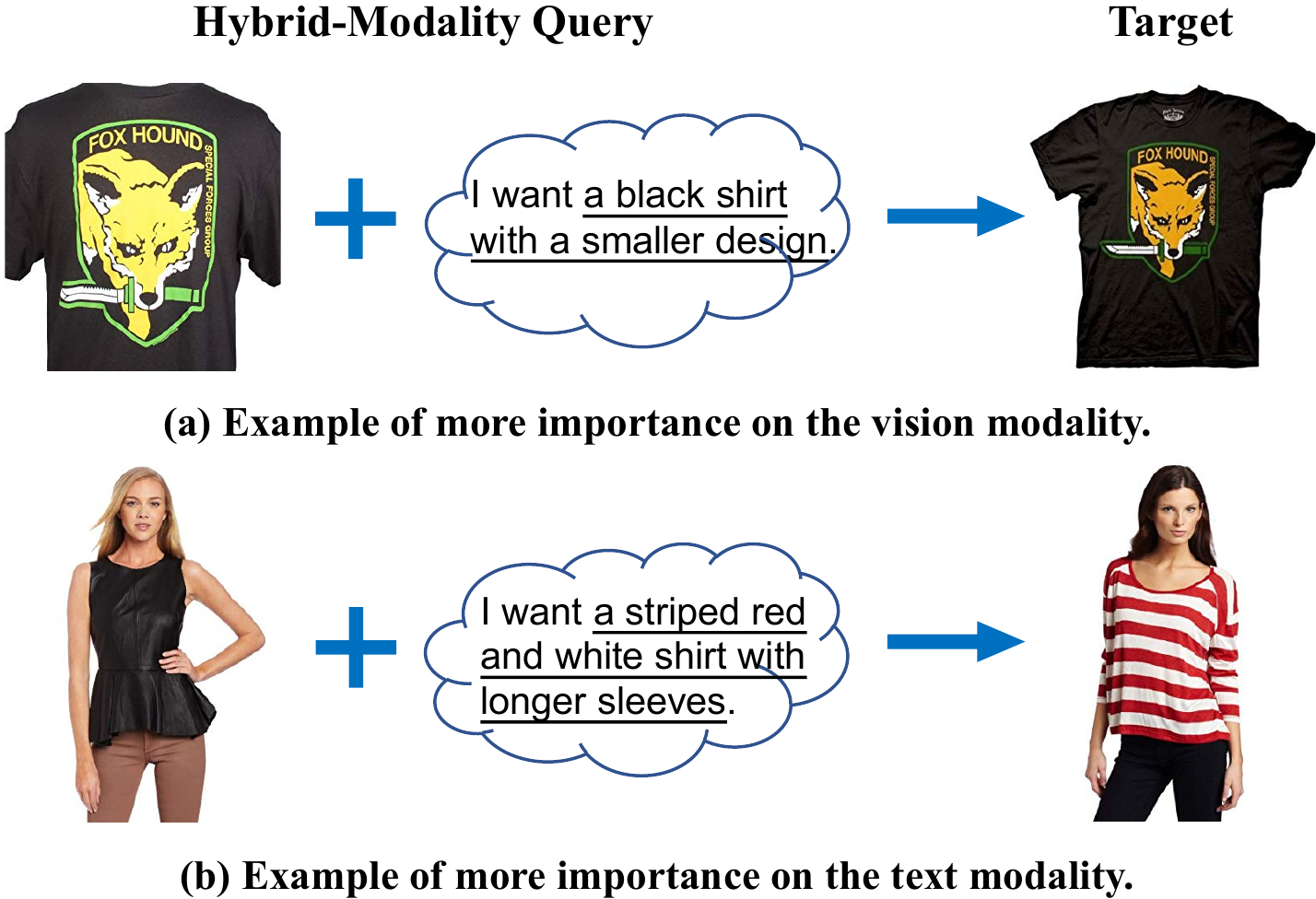}
  \caption{Illustration of image retrieval with hybrid-modality queries. The importance of each modality in the hybrid-modality query varies for different retrieval intentions.}
  \label{fig:con2}
\end{figure}

With the continuous emergence of a large amount of multimedia data on the Internet, multimedia retrieval has become a basic technology to meet the needs of users for information access, such as searching for a product image similar to the query picture \cite{Liu2016DeepFashion}, or retrieving images via descriptive text queries \cite{wang2016image-text,faghri2018vsepp,gao2020fashionbert}.
The query can be in the same modality as the retrieval target or in a different modality as in cross-modal retrieval \cite{wang2016image-text,faghri2018vsepp,Sangkloy2016sketchy}.
However, in either case, the query from a single modality is limited to express complex search intentions, such as searching an image similar to a reference image but with some modifications described in the text. In such case, the search intention is expressed in a more complex query format with hybrid-modality that involves both vision and text modalities as illustrated in Figure~\ref{fig:con2}.
This is especially common in the fashion e-commerce scenario, where users tend to search for a similar fashion product when seeing a reference picture, but at the same time wanting certain changes over the reference, such as changing the color, length, style, etc. 
In addition, composing text and image for image retrieval also enables multi-round retrieval with text feedback \cite{wu2021fashioniq, Guo2018Shoes, Yuan2021MultiturnFashionIQ}, which has a great application potential in the dialog-based interactive retrieval scenario \cite{Guo2018Shoes}.
In this paper, we focus on the image retrieval with hybrid-modality queries problem, also known as the composing text and image for image retrieval (CTI-IR) task \cite{vo2019tirg}.

Previous approaches \cite{vo2019tirg, Lee2021CosMo, chen2020val, wen2021clvcnet, Hosseinzadeh2020LBFCVPR2020,Kim2021DCNet,gu2021DATIR,yang2021JPM} for this task can be categorized into two types.
The first type of works \cite{vo2019tirg, Lee2021CosMo, chen2020val, wen2021clvcnet, Hosseinzadeh2020LBFCVPR2020} mainly focus on designing complex components for the multi-modal fusion between text and image queries.
For example, Vo et al. \cite{vo2019tirg} propose a TIRG model with gated residual connection to modify partial image regions with text guidance and keep others unchanged.
Wen et al. \cite{wen2021clvcnet} propose to combine local-wise and global-wise composition modules for both local and global modification demands.
The second type of works \cite{Kim2021DCNet,gu2021DATIR,yang2021JPM} focus on enhancing the semantic embedding space by combining the \textit{image\&text}-to-\textit{image} matching and \textit{image\&image}-to-\textit{text} matching with multi-task learning.
For example, Yang et al. \cite{yang2021JPM} propose an auxiliary module to align the difference between reference and target images with the modification text by joint prediction.

However, due to the scarcity of supervised data which needs to be in the triplet format as <reference-image, modification-text, target-image>, and the complexity of CTI-IR task which requires both the semantic space learning for target retrieval and cross-modal fusion between hybrid-modality queries, it is hard to effectively learn the complex knowledge together and thus the existing methods in both two types achieve marginally satisfactory retrieval results, with only about 30\% of queries retrieving the correct image in the top-10 rank.
Based on our analysis, we consider three challenges to be tackled in this task.
First, similar to other image retrieval tasks, a semantic embedding space needs to be learned, in which instances with similar semantics even in different modalities are closer, while irrelevant ones are away from each other.
Second, since the CTI-IR task mainly focuses on the fashion domain, the domain-specific triplet training examples are rather scarce, making it difficult to learn effective multi-modal fusion for hybrid-modality queries.
Last but not least, the individual modality contribution in the hybrid-modality query varies for different retrieval intentions.
As shown in Figure~\ref{fig:con2}, the vision modality is more important than text when the users expect to search for a fashion product much similar to the reference product image, while the text modality is more important when more detailed descriptions for the target image are provided.
Therefore, the model needs to learn how to adaptively assign different importance on the image and text queries according to their characteristics.

\begin{figure}
  \centering
  \includegraphics[width=\linewidth]{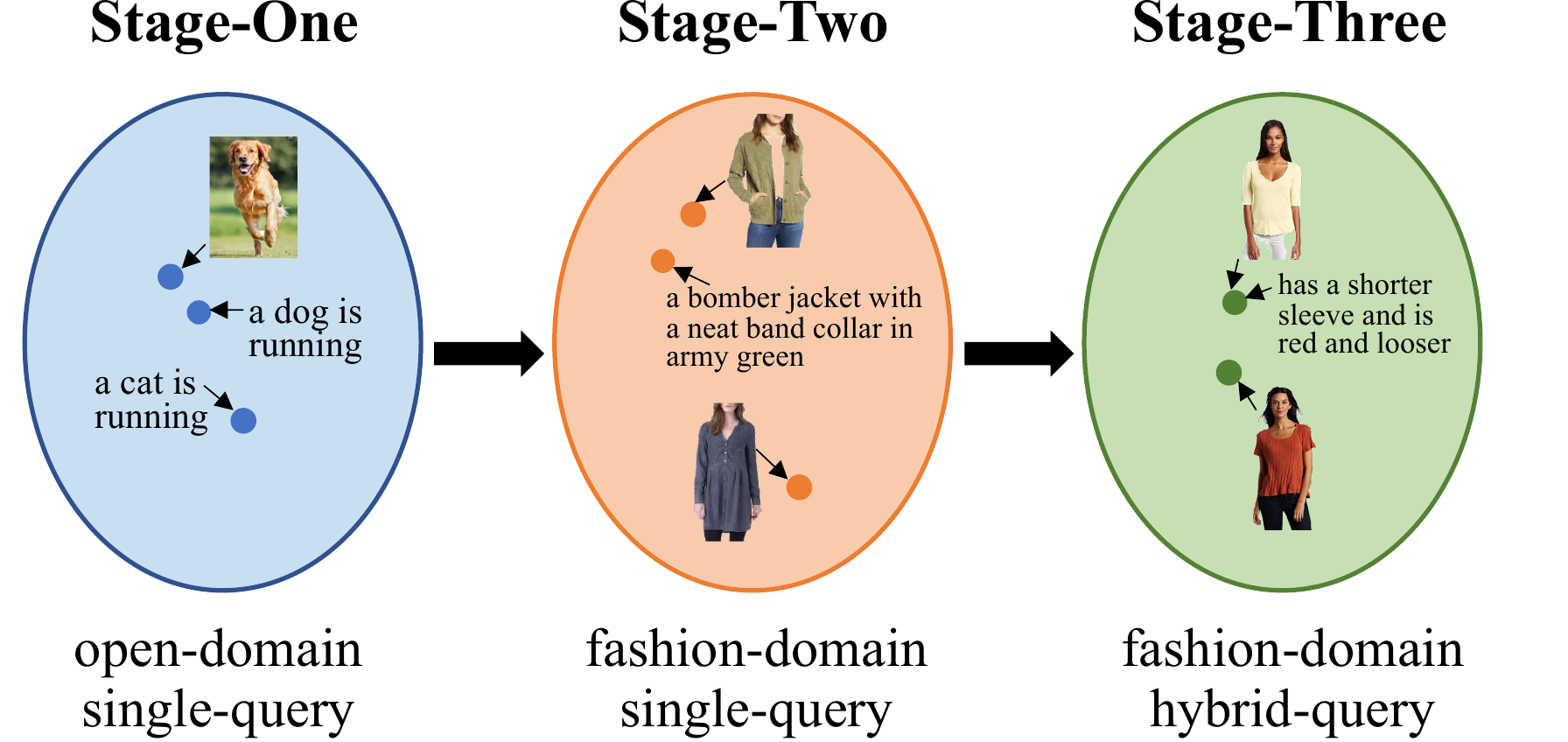}
  \caption{Concept illustration of the proposed three-stage progressive learning for the CTI-IR task.}
  \label{fig:con1}
\end{figure}

Considering that the above three challenges are difficult to solve perfectly together, in this paper, we decompose the CTI-IR task into a three-stage learning problem to acquire the knowledge step by step progressively.
Inspired by the prominent success of CLIP \cite{Radford2021CLIP} (Contrastive Language-Image Pre-training) on the open-domain image-text cross-modal retrieval due to its pre-training on large-scale image and text pairs, we propose to leverage the open-domain semantic joint embedding space learned by CLIP.
As shown in Figure~\ref{fig:con1}, we first utilize the capability of CLIP on the open-domain single-query image retrieval, which can project semantically relevant images and texts closer, while the irrelevant instances farther away.
Then, we transfer the learned knowledge from open-domain to the fashion domain by further pre-training the model with fashion-related pre-training tasks.
Finally, we adapt the model from single-query to the CTI-IR task with hybrid-modality query on the limited <reference-image, modification-text, target-image> triplet data.
With the three-stage progressive learning, our model can easily exploit rich resources in other data format and from other domains, and learn the complex knowledge progressively for the CTI-IR task.
To dynamically fuse the image and text queries for different retrieval intentions, we propose an adaptive weighting strategy learned by self-supervised pseudo labels to dynamically assign different focuses on the image and text queries.
Extensive experiments on the Fashion-IQ \cite{wu2021fashioniq} and Shoes \cite{Guo2018Shoes} benchmark datasets show that our simple but effective model significantly outperforms existing methods.
Additionally, qualitative visualization shows that our model can indeed adaptively focus more on the image or text query to better retrieve the target image.

The main contributions of this work are as follows:
\begin{itemize}
    \item We decompose the CTI-IR task into a three-stage learning problem to acquire the complex knowledge progressively, and fully exploit the open-domain and open-format resources to alleviate the data scarcity of fashion-domain triplet data.
    \item We analyze the roles that image and text queries play for different retrieval intentions, and propose an adaptive multi-modal weighting strategy to dynamically compose image and text queries.
    \item Our model significantly outperforms state-of-the-art methods on Fashion-IQ and Shoes benchmark datasets, with the mean of Recall@K improved by 24.9\% and 9.5\% respectively.
\end{itemize}

\section{Related Work}
\subsection{Image Retrieval}
Image retrieval is an important research problem in both computer vision and information retrieval, which searches for an image from a large-scale database given a query.
The query can be an image \cite{Liu2016DeepFashion}, or in other modalities such as the descriptive text \cite{wang2016image-text,faghri2018vsepp} or a sketch \cite{Sangkloy2016sketchy,huang2017sketch}, which is called cross-modal image retrieval.
The mainstream approach is to learn a joint semantic embedding space, in which the query and retrieval candidates with similar semantics will be projected closer, and otherwise farther away.
Although excellent results have been achieved, the single-query has a poor performance in expressing complex retrieval intentions.
Recently, Vo et al. \cite{vo2019tirg} propose composing text and image for the image retrieval (CTI-IR), where the image query stands for a reference and the text query describes some modification requests of the reference image.
Subsequently, more and more works focus on this CTI-IR task.
The previous works \cite{vo2019tirg, Lee2021CosMo, chen2020val, wen2021clvcnet, Hosseinzadeh2020LBFCVPR2020, Kim2021DCNet, yang2021JPM, gu2021DATIR, Chen2020JSVMECCV, zhang2020jointCTIIR} can be categorized into two types.
The first type of works \cite{vo2019tirg, Lee2021CosMo, chen2020val, wen2021clvcnet, Hosseinzadeh2020LBFCVPR2020} mainly focus on the multi-modal fusion between image and text queries.
The TIRG \cite{vo2019tirg} model fuses the reference image features and the text representation with a gated residual connection to modify parts of the image and keep others unchanged.
The VAL \cite{chen2020val} model exploits fine-grained local image feature maps and fuses them with the text representation via attention mechanism.
The recently proposed CLVC-Net \cite{wen2021clvcnet} combines local-wise and global-wise image-text compositions for various image modification demands.
The second type of works \cite{Kim2021DCNet, yang2021JPM, gu2021DATIR} propose to train the model with both \textit{image\&text}-to-\textit{image} matching and \textit{image\&image}-to-\textit{text} matching objectives to improve the semantic space learning.
However, both types of approaches attempt to simultaneously learn the semantic embedding space for target retrieval and the multi-modal fusion between hybrid-modality queries, which is hard especially when the supervised data is limited.
Moreover, the previous works also do not consider that the importance of each modality in hybrid-modality queries is different for different retrieval intentions.
In this work, we propose a progressive learning strategy to fully exploit the single-query image retrieval knowledge in open-domain and fashion-domain, and adaptively adjust the roles of image and text queries for the CTI-IR task.

\subsection{Vision-Language Pre-training}
Inspired by the success of large-scale transformer-based pre-training models \cite{Devlin2019BERT,radford2019gpt2} in the field of Natural Language Processing, similar attempts are also conducted in vision and language research. 
There are mainly two types of architecture choices for Vision-Language Pre-training (VLP) models, including single-stream and two-stream architectures.
The single-stream VLP models \cite{li2019visualbert,li2020oscar,chen2019uniter,li2020unicodervl} encode image features and text embeddings with a single transformer encoder, which explores the multi-modal interactions at an earlier stage.
The two-stream VLP models \cite{lu2019vilbert,tan2019lxmert} however first encode the two modalities separately, and then combine the features from the two branches with a cross transformer encoder for the multi-modal fusion.
Although impressive results have been achieved in many vision-and-language downstream tasks, e.g. VQA \cite{antol2015vqa,anderson2018butd,gao2019vqa} and image captioning \cite{vinyals15showtell,anderson2018butd,huang2019aoanet}, both the single-stream and two-stream methods suffer from the low retrieval efficiency problem due to the cross fusion encoder, and thus are inappropriate for the image retrieval task.
Radford et al. \cite{Radford2021CLIP} propose a Contrastive Language-Image Pre-training (CLIP) model with only two independent encoders and achieve superior retrieval results in open-domain image-text retrieval task.
Therefore, we fully exploit the advantages of CLIP and propose to transfer the semantic alignment between image and text from open-domain to the fashion-domain.

There are also some fashion-domain VLP models related to our work, such as FashionBERT \cite{gao2020fashionbert} and Kaleido-BERT \cite{Zhuge2021kaleido}.
However, both of them adopt the single-stream architecture and thus have low efficiency for the retrieval inference.
Additionally, they only focus on the single-query image-text retrieval tasks (\textit{text}-to-\textit{image} and \textit{image}-to-\textit{text}) without considering the hybrid-modality query scenario that composes both vision and text modalities.
In this work, we conduct the fashion-domain pre-training with two different pre-training tasks based on CLIP for the image retrieval with hybrid-modality queries.

\begin{figure*}
  \centering
  \includegraphics[width=\linewidth]{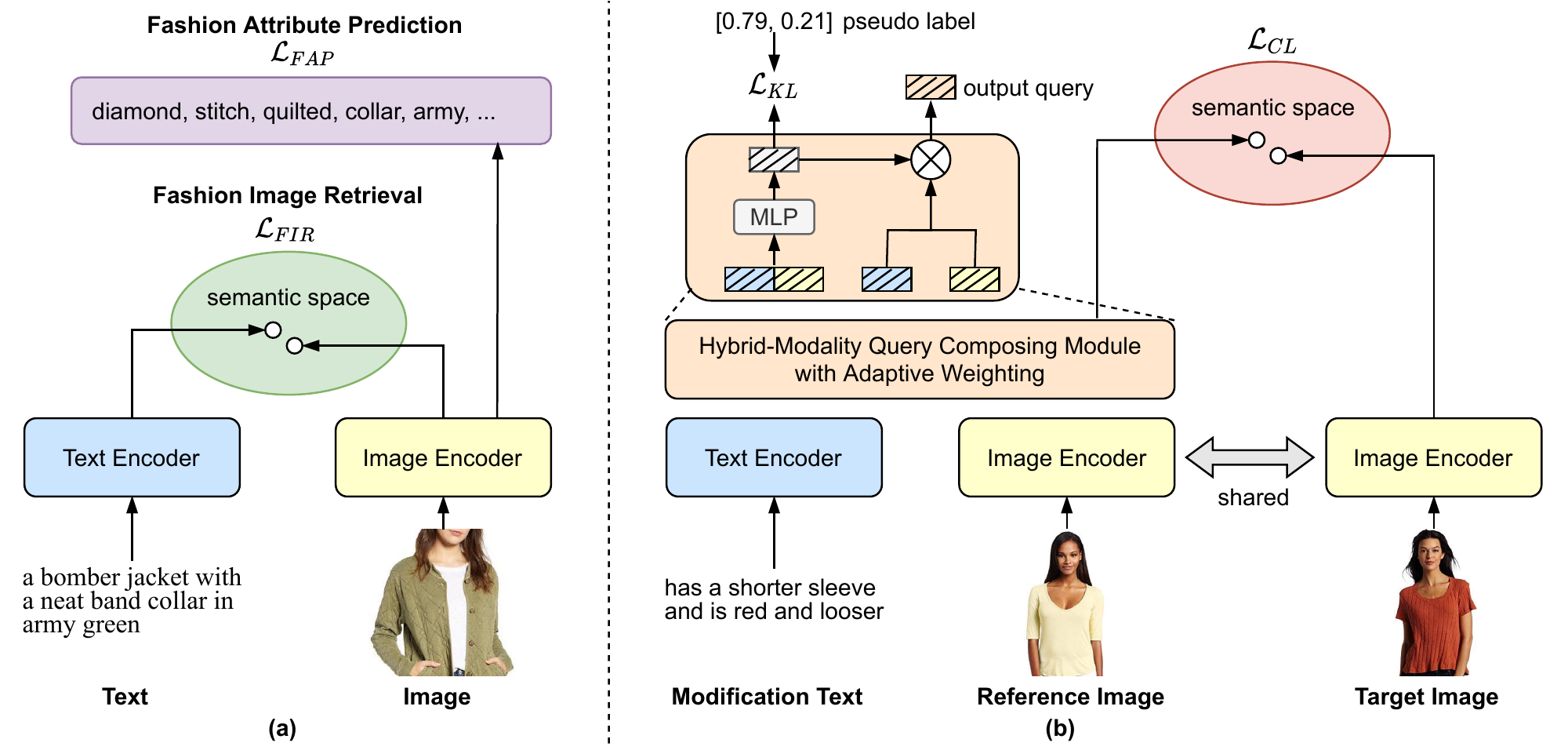}
  \caption{Illustration of our model with progressive learning for the CTI-IR task, (a) shows our model pre-trained with fashion-related tasks in Stage-Two, (b) shows our model adapted for the hybrid-modality image retrieval in Stage-Three.}
  \label{fig:model}
\end{figure*}

\section{Methodology}
Given a hybrid-modality query $(I_r, T)$ composing a reference image $I_r$ and a descriptive text $T$, the goal of CTI-IR task is to retrieve the corresponding target image $I_t$ from a large candidate image set $\Theta_I$.
The model is optimized to pull the embedding of $\mathcal{F}(I_r, T)$ closer to the embedding of $I_t$, while push it away from the embedding of $I_{\setminus t} \in \Theta_I$ in the semantic joint embedding space, where $\mathcal{F}(\cdot)$ denotes the hybrid-modality query composing module.
Since both the hybrid-modality query composing module $\mathcal{F}(\cdot)$ and the semantic joint embedding space have high impact on the final retrieval result, directly optimizing them together from scratch can achieve sub-optimal result.
Therefore, we decompose the task into a three-stage learning problem, which progressively learns from open-domain to fashion-domain, and then from single-query to hybrid-modality query for image retrieval.
Specifically, we leverage CLIP \cite{Radford2021CLIP} as the source of our first-stage open-domain image retrieval knowledge to initialize the model, and then pre-train it on the fashion-domain vision-language data via two fashion-related tasks to transfer the knowledge into the fashion domain.
Finally, we further optimize the model with self-supervised adaptive query composing on the limited $<I_r, T, I_t>$ triplets for the CTI-IR task.

\subsection{Model Architecture}
\label{sec:model}
Due to the characteristics of CTI-IR task, which takes a reference image and a descriptive text as the query input to search for relevant target images, our model contains two image encoders and one text encoder.
The two image encoders share parameters, with one for the reference image encoding and the other for the target image encoding.
A hybrid-modality query composing module $\mathcal{F}(\cdot)$ is applied on the reference image encoder and the text encoder to combine the multi-modal queries for the target image retrieval.
Following the CLIP \cite{Radford2021CLIP} setting, we use the 12-layer BERT \cite{Devlin2019BERT} as the text encoder, ResNet-50 \cite{He2016resnet} or ViT-B/32 \cite{Do2021ViT} as the image encoder for our base and large model respectively.

\subsection{Multi-Stage Progressive Learning}
\label{sec:con1}
To learn the semantic embedding and multi-modal fusion knowledge progressively and fully leverage the open-domain and open-format resources, we propose to optimize our CTI-IR model through three stages.
It first leverages the open-domain image retrieval capability from pre-trained vision-language models, then learns the domain-specific image retrieval capability through designed domain-specific pre-training tasks, and finally masters the capability for image retrieval with hybrid-modality queries through task-specific fine-tuning.

\noindent \textbf{Stage-One:} 
We first initialize the image encoder and text encoder using the CLIP backbone, so that our model has the initial ability to encode open-domain (image, text) pairs into closely matching vectors.

\noindent \textbf{Stage-Two:} 
Since CLIP is pre-trained on the open-domain (image, text) corpus, it performs poorly in the fashion scenario. Therefore, we further pre-train the model on fashion-domain vision-language data with two pre-training tasks: Fashion Image Retrieval (FIR) and Fashion Attribute Prediction (FAP), as shown in the Figure~\ref{fig:model} (a).

We train the FIR task on the fashion-domain image-text pairs $(I_f^{(i)}, T_f^{(i)}) \in \mathcal{D}_1$, where $I_f^{(i)}$ is a fashion product image and $T_f^{(i)}$ is the description of the corresponding fashion product.
We feed the image $I_f^{(i)}$ and text $T_f^{(i)}$ into the image encoder and text encoder respectively, and optimize their global encoding vectors $z_I^{(i)}$ and $z_T^{(i)}$ via contrastive learning with bi-directional InfoNCE \cite{oord2018infoNCE} loss as follows:
\begin{equation}
\begin{aligned}
    \mathcal{L}_{FIR} = \frac{1}{B}\sum_i^B -\log\frac{\exp(s(z_I^{(i)}, z_T^{(i)})/ \tau_1)}{\sum_{j=1}^B \exp(s(z_I^{(i)}, z_T^{(j)})/ \tau_1)} \\ +\frac{1}{B}\sum_i^B -\log\frac{\exp(s(z_I^{(i)}, z_T^{(i)})/ \tau_1)}{\sum_{j=1}^B \exp(s(z_I^{(j)}, z_T^{(i)})/ \tau_1)},
\end{aligned}
\end{equation}
where $B$ is the mini-batch size, $s(\cdot)$ denotes the cosine similarity function and $\tau_1$ is the learnable temperature. The FIR task helps to learn the fashion-domain semantic embedding space for fashion image retrieval.

To further enhance the image encoder for capturing more local details of the fashion product which is important for the CTI-IR task, we propose a FAP task to predict the attributes of the fashion product image with an additional multi-label classification head on the image encoder.
We train the FAP task on the fashion product images with multiple annotated attribute labels $(I_f^{(i)}, A^{(i)}) \in \mathcal{D}_2$, where $A^{(i)}=\{a_1^{(i)},a_2^{(i)},\cdots, a_n^{(i)}\}$ refers to the multiple attributes.
We optimize the FAP task with cross entropy loss as follows:
\begin{eqnarray}
    & p^{(i)} = \sigma (W \cdot z_I^{(i)} + b), \\
    & \mathcal{L}_{FAP} = -\sum\limits_i\sum\limits_k(a_k^{(i)}\log p_k^{(i)}+(1-a_k^{(i)})\log (1-p_k^{(i)})),
\end{eqnarray}
where $\sigma$ denotes the sigmoid function, $z_I^{(i)}$ is the encoded vector of $I_f^{(i)}$, $W$ and $b$ are parameters of the attribute classification layer.

We iteratively pre-train the model with the FIR and FAP tasks at a 1:1 ratio to transfer the knowledge of CLIP from the open-domain to the fashion-domain image retrieval.

\noindent \textbf{Stage-Three:} 
Through the first two stages of training, the model has the capability of single-query image retrieval in the fashion domain. However, it cannot yet handle the image retrieval with hybrid-modality queries. Therefore, in the third stage, we adapt the model to the CTI-IR task on the downstream $<I_r^{(i)}, T^{(i)}, I_t^{(i)}> \in \mathcal{D}_3$ triplet data as shown in Figure~\ref{fig:model} (b).
We equip the model with a hybrid-modality query composing module $\mathcal{F}(\cdot)$ and further optimize the model end to end via contrastive learning for the composed query feature $\mathcal{F}(z_{I_r}^{(i)}, z_T^{(i)})$ and the target image feature $z_{I_t}^{(i)}$ as follows:
\begin{equation}
    z_Q^{(i)} = \mathcal{F}(z_{I_r}^{(i)}, z_T^{(i)}),
\end{equation}
\begin{equation}
    \mathcal{L}_{CL} = \frac{1}{B}\sum_i^B - \log\frac{\exp(s(z_Q^{(i)},  z_{I_t}^{(i)})/ \tau_2)}{\sum_{j=1}^B \exp(s(z_Q^{(i)}, z_{I_t}^{(j)})/ \tau_2)},
\end{equation}
where $z_{I_r}^{(i)}$, $z_T^{(i)}$ and $z_{I_t}^{(i)}$ are the global encoded vectors from image and text encoders for $I_r^{(i)}$, $T^{(i)}$ and $I_t^{(i)}$ respectively. $s(\cdot)$ denotes the cosine similarity function, $\tau_2$ is the learnable temperature and $B$ is the size of mini-batch.

Through the multi-stage learning and specific pre-training tasks designed for each stage, the model can not only progressively learn the knowledge required by the CTI-IR task, but also make full use of extra data to facilitate learning, such as $\mathcal{D}_1$ and $\mathcal{D}_2$, even though they are not in the triplet format required by the CTI-IR task.

\subsection{Self-Supervised Query Adaptive Weighting}
\label{sec:con2}
The query composing function $\mathcal{F}(\cdot)$ that fuses the multi-modal queries is important for the CTI-IR task.
Although the previous works \cite{Lee2021CosMo, chen2020val, wen2021clvcnet} design complex multi-modal fusion approaches with attention mechanism, the importance weights of image and text queries are not explicitly explored in these methods, which can be different for different retrieval intentions.
Therefore, we propose a query adaptive weighting strategy to dynamically assign different weights for image and text queries based on their importance to the retrieval intention via self-supervision.

Specifically, we concatenate the encoded feature vectors of $I_r^{(i)}$ and $T^{(i)}$ into a one-layer MLP to get the importance weights for image and text queries.
Then, the predicted weights are multiplied to the vectors to get the final composed query feature, which is expressed as follows:
\begin{eqnarray}
    & \widetilde{w}^{(i)} = \text{MLP}([z_{I_r}^{(i)}, z_T^{(i)}]), \quad \widetilde{w}^{(i)} \in \mathbb{R}^2, \\
    & z_Q^{(i)} = \widetilde{w}^{(i)} \cdot [(z_{I_r}^{(i)})^\mathrm{T}, (z_T^{(i)})^\mathrm{T}]^\mathrm{T}.
\end{eqnarray}
To make the model learn effective modality weights $\widetilde{w}^{(i)}$, we automatically generate pseudo labels for the weight matrix supervision.
To evaluate the roles of image and text queries played in different retrieval cases, we pre-train three models: \emph{image-only} model, \emph{text-only} model and \emph{fusion} model.
The \emph{image-only} model is trained to retrieve the target image $I_t^{(i)}$ only based on the reference image $I_r^{(i)}$, while the \emph{text-only} model is to retrieve $I_t^{(i)}$ only based on the descriptive text $T^{(i)}$.
The \emph{fusion} model is similar to our base model except that the composing module $\mathcal{F}(\cdot)$ is just the mean pooling.

We use the three models to infer the rank of ground-truth target image $I_t^{(i)}$ respectively, and produce three ranks ($r_{\text{image}}^{(i)}$, $r_{\text{text}}^{(i)}$, $r_{\text{fuse}}^{(i)}$) for each triplet example.
The model who ranks the ground-truth target image higher (the rank number is smaller) means that the model performs better for this retrieval case.
Therefore, the performances of \emph{image-only} and \emph{text-only} model can reflect the importance of the image and text queries.
We denote the retrieval performance of each model for the $i$-th example as the inverse of normalized ranks as follows:
\begin{eqnarray}
    & s_{I_r}^{(i)} = (r_{\text{image}}^{(i)} / N)^{-1}, \quad s_{T}^{(i)} = (r_{\text{text}}^{(i)} / N)^{-1}, \\
    & s_f^{(i)} = (r_{\text{fuse}}^{(i)} / N)^{-1},
\end{eqnarray}
where $N$ is the number of total candidates.
Since the retrieval difficulty of different queries can vary greatly, the absolute disparity between $s_{I_r}^{(i)}$ and $s_{T}^{(i)}$ is also influenced by the overall retrieval performance and thus cannot reflect the relative divergence.
Therefore, we further divide them by the score $s_f^{(i)}$ from the \emph{fusion} model.
Finally, we acquire the pseudo importance weight labels as follows:
\begin{equation}
    w^{(i)} = softmax(\tau_3 \cdot [s_{I_r}^{(i)} / s_f^{(i)}, s_{T}^{(i)} / s_f^{(i)}]),
\end{equation}
where $N$ is the number of total candidates, and $\tau_3$ is the temperature hyper-parameter.

With such pseudo labels, we optimize the learning of the hybrid-modality query composing via KL divergence loss as follows:
\begin{equation}
    \mathcal{L}_{KL} = \sum\limits_i D_{KL}(w^{(i)}||\widetilde{w}^{(i)}).
\end{equation}
Finally, we combine $\mathcal{L}_{KL}$ and $\mathcal{L}_{CL}$ with a hyper-parameter $\lambda$ as the total loss for stage-three:
\begin{equation}
\label{eq:total_loss}
    \mathcal{L}_{total} = \mathcal{L}_{CL}+ \lambda\mathcal{L}_{KL}.
\end{equation}

\vspace{3pt}
\section{Experiments}
In this section, we evaluate our model on image retrieval with hybrid-modality queries.
There are two versions of our model including the base model and the large model, corresponding to using the ResNet-50 and ViT-B/32 image encoder backbones respectively.

\subsection{Datasets}
Thanks to our multi-stage progressive learning strategy, we can easily leverage other available fashion domain data to alleviate the scarcity of fashion-domain triplet examples in the CTI-IR task.
We conduct the stage-two fashion-domain pre-training on three datasets, including FACAD \cite{Yang2020FACAD}, FashionGen \cite{rostamzadeh2018fashiongen} and DeepFashion \cite{Liu2016DeepFashion}.
The FACAD and FashionGen datasets are used to train the FIR (Fashion Image Retrieval) task, and the DeepFashion dataset is used for the FAP (Fashion Attribute Prediction) task.

\textbf{FACAD} \cite{Yang2020FACAD} contains 126,753 English fashion product descriptions, with each product description accompanied by average 6 \textasciitilde 7 product images of different colors and poses.
We remove the redundant examples and those with too long or short descriptions, finally acquiring 853,503 image-text pairs in total.

\textbf{FashionGen} \cite{rostamzadeh2018fashiongen} is widely used in fashion-domain pre-training works, such as the FashionBERT \cite{gao2020fashionbert} and Kaleido-BERT \cite{Zhuge2021kaleido}. It contains 67,666 fashion product descriptions with 1 \textasciitilde 6 product images from different angles. Following the FashionBERT \cite{gao2020fashionbert}, we use 260,480 image-text pairs for training, and 35,528 for test.

\textbf{DeepFashion} \cite{Liu2016DeepFashion} is a large-scale clothes dataset which are annotated with categories, attributes, landmarks, etc. 
In this work, we only use the image and attribute data, which contains 209,222 images with average 3.3 attribute labels per image.
The total number of attribute classes is 1,000.

\begin{table}[t]
\centering
\caption{The statistics of different retrieval candidate sets for evaluation in the Fashion-IQ dataset.}
\label{tab:data_split}
\begin{tabular}{l|ccc}
\toprule
Split & Dress & Shirt & Tops\&Tees \\
\midrule
Original split \cite{wu2021fashioniq} & 3,817 & 6,346 & 5,373 \\
VAL split \cite{chen2020val} & 2,628 & 3,089 & 2,902 \\
\bottomrule
\end{tabular}
\end{table}

\begin{table*}[t]
\centering
\caption{Comparisons with the state-of-the-art methods for image retrieval with hybrid-modality queries on Fashion-IQ dataset. We report the results of our base and large models on the two retrieval candidate splits. ``$\dagger$'' denotes the result is significantly better than the best baseline at $p \textless 0.05$. The best results are in bold and the second best are underlined.}
\label{tab:fashioniq}
\begin{tabular}{c|l|cccccc|ccc}
\toprule
\multirow{2}{*}{Split} & \multirow{2}{*}{Method} & \multicolumn{2}{c}{Dress} & \multicolumn{2}{c}{Shirt} & \multicolumn{2}{c}{Tops\&Tees} & \multicolumn{3}{|c}{Avg} \\
&& R@10 & R@50 & R@10 & R@50 & R@10 & R@50 & R@10 & R@50 & Rmean \\
\midrule
\multirow{10}{*}{VAL split \cite{chen2020val}} & Film \cite{Perez2018Film} & 14.23 & 33.34 & 15.04 & 34.09 & 17.30 & 37.68 & 15.52 & 35.04 & 25.28\\
&TIRG \cite{vo2019tirg} & 14.87 & 34.66 & 18.26 & 37.89 & 19.08 & 39.62 & 17.40 & 37.39 & 27.40\\
&Relationship \cite{Santoro2017Relationship} & 15.44 & 38.08 & 18.33 & 38.63 & 21.10 & 44.77 & 18.29 & 40.49 & 29.39\\
&VAL \cite{chen2020val} & 22.53 & 44.00 & 22.38 & 44.15 & 27.53 & 51.68 & 24.15 & 46.61 & 35.38\\
&DATIR \cite{gu2021DATIR} & 21.90 & 43.80 & 21.90 & 43.70 & 27.20 & 51.60 & 23.70 & 46.40 & 35.05\\
&JPM \cite{yang2021JPM} & 21.38 & 45.15 & 22.81 & 45.18 & 27.78 & 51.70 & 23.99 & 47.34 & 35.67\\
&CoSMo \cite{Lee2021CosMo} & 25.64 & 50.30 & 24.90 & 49.18 & 29.21 & 57.46 & 26.58 & 52.31 & 39.45\\
&CLVC-Net \cite{wen2021clvcnet} & 29.85 & 56.47 & 28.75 & 54.76 & 33.50 & 64.00 & 30.70 & 58.41 & 44.56\\
&Ours-base & $\underline{33.22}^\dagger$ & $\underline{59.99}^\dagger$ & $\underline{46.17}^\dagger$ & $\underline{68.79}^\dagger$ & $\underline{46.46}^\dagger$ & $\underline{73.84}^\dagger$ & $\underline{41.98}^\dagger$ & $\underline{67.54}^\dagger$ & $\underline{54.76}^\dagger$\\
&\textbf{Ours-large} & $\textbf{38.18}^\dagger$ & $\textbf{64.50}^\dagger$ & $\textbf{48.63}^\dagger$ & $\textbf{71.54}^\dagger$ & $\textbf{52.32}^\dagger$ & $\textbf{76.90}^\dagger$ & $\textbf{46.37}^\dagger$ & $\textbf{70.98}^\dagger$ & $\textbf{58.68}^\dagger$\\
\midrule
\multirow{5}{*}{Original split \cite{wu2021fashioniq}} &TIRG \cite{vo2019tirg} & 14.13 & 34.61 & 13.10 & 30.91 & 14.79 & 34.37 & 14.01 & 33.30 & 23.66\\
&CoSMo \cite{Lee2021CosMo} & 21.39 & 44.45 & 16.90 & 37.49 & 21.32 & 46.02 & 19.87 & 42.62 & 31.25\\
&DCNet \cite{Kim2021DCNet} & 28.95 & \underline{56.07} & 23.95 & 47.30 & 30.44 & 58.29 & 27.78 & 53.89 & 40.84\\
&Ours-base & \underline{29.00} & 53.94 & $\underline{35.43}^\dagger$ & $\underline{58.88}^\dagger$ & $\underline{39.16}^\dagger$ & $\underline{64.56}^\dagger$ & $\underline{34.53}^\dagger$ & $\underline{59.13}^\dagger$ & $\underline{46.83}^\dagger$ \\
&\textbf{Ours-large} & $\textbf{33.60}^\dagger$ & $\textbf{58.90}^\dagger$ & $\textbf{39.45}^\dagger$ & $\textbf{61.78}^\dagger$ & $\textbf{43.96}^\dagger$ & $\textbf{68.33}^\dagger$ & $\textbf{39.02}^\dagger$ & $\textbf{63.00}^\dagger$ & $\textbf{51.01}^\dagger$\\
\bottomrule
\end{tabular}
\end{table*}

We evaluate our model on two benchmark CTI-IR datasets, including Fashion-IQ \cite{wu2021fashioniq} and Shoes \cite{Guo2018Shoes}.

\textbf{Fashion-IQ} \cite{wu2021fashioniq} is a natural language based interactive fashion image retrieval dataset involving three common product categories: Dress, Shirt, and Tops\&Tees.
It contains 18,000 <reference-image, modification-text, target-image> triplets for training, 6,016 for validation and 6,118 for the test. Since the ground-truth of the test set is reserved for the challenge\footnote{https://sites.google.com/view/cvcreative2020/fashion-iq}, we follow previous works \cite{Lee2021CosMo, chen2020val, wen2021clvcnet} to evaluate our model on the validation set.
The original Fashion-IQ dataset provides retrieval candidate image set for each product category, with 3,817, 6,346 and 5,373 images respectively.
However, the VAL \cite{chen2020val} model is evaluated with the self-constructed candidate set, which is the union of the reference and target images with redundancy removed. It is smaller than the original candidate set as shown in the Table~\ref{tab:data_split}.
For a fair comparison with other methods, we report the retrieval results on both candidate sets.

\textbf{Shoes} \cite{Guo2018Shoes} is another natural language based interactive fashion image retrieval dataset, where the images are crawled from like.com website \cite{Berg2010ShoesAttr}. The fashion products are all about shoes.
There are 10,000 images in the training set and 4,658 in the test set, with 10,751 triplets annotated in total.
All the images in the test set are used as retrieval candidates for evaluation.

\subsection{Evaluation Metrics}
Following previous works \cite{vo2019tirg, chen2020val, wen2021clvcnet}, we evaluate the model performance with Recall@K (R@K), which refers to the ratio of queries that correctly retrieve the ground-truth target image in the top-K ranking list.
The value of K is 10 and 50 on Fashion-IQ dataset, and 1, 10, 50 on Shoes dataset.
We also report the mean of all R@K values as Rmean for the overall retrieval performance.

\begin{table}[t]
\centering
\caption{Comparison with the state-of-the-art models on the Shoes dataset. ``$\dagger$'' denotes the result is significantly better than the best baseline at a significance level $p \textless 0.05$. The best results are in bold and the second best are underlined.}
\label{tab:shoes}
\begin{tabular}{lccc|c}
\toprule
Method & R@1 & R@10 & R@50 & Rmean\\
\midrule
Film \cite{Perez2018Film} & 10.19 & 38.89 & 68.30 & 39.13\\
TIRG \cite{vo2019tirg} & 12.60 & 45.45 & 69.39 & 42.48\\
Relationship \cite{Santoro2017Relationship} & 12.31 & 45.10 & 71.45 & 42.95\\
VAL \cite{chen2020val} & 17.18 & 51.52 & 75.83 & 48.18\\
DATIR \cite{gu2021DATIR} & 17.20 & 51.10 & 75.60 & 47.97\\
CoSMo \cite{Lee2021CosMo} & 16.72 & 48.36 & 75.64 & 46.91\\
DCNet \cite{Kim2021DCNet} & - & 53.82 & 79.33 & -\\
CLVC-Net \cite{wen2021clvcnet} & 17.64 & 54.39 & 79.47 & 50.50\\
\midrule
Ours-base & $\underline{19.53}^\dagger$ & $\underline{55.65}^\dagger$ & $\underline{80.58}^\dagger$ & $\underline{51.92}^\dagger$\\
\textbf{Ours-large} & $\textbf{22.88}^\dagger$ & $\textbf{58.83}^\dagger$ & $\textbf{84.16}^\dagger$ & $\textbf{55.29}^\dagger$\\
\bottomrule
\end{tabular}
\end{table}

\begin{table*}[t]
\centering
\caption{Ablation study of our base model with different training stages in progressive learning on the Fashion-IQ dataset.}
\label{tab:abalation_progress}
\begin{tabular}{c|ccc|cccccc|ccc}
\toprule
\multirow{2}{*}{Row} & \multicolumn{3}{c|}{Progressive Learning}  & \multicolumn{2}{c}{Dress} & \multicolumn{2}{c}{Shirt} & \multicolumn{2}{c|}{Tops\&Tees} & \multicolumn{3}{c}{Avg} \\
&Stage 1 & Stage 2 & Stage 3 & R@10 & R@50 & R@10 & R@50 & R@10 & R@50 & R@10 & R@50 & Rmean \\
\midrule
1 & \checkmark & & & 5.75 & 13.04 & 11.73 & 21.98 & 11.22 & 21.93 & 9.57 & 18.98 & 14.27 \\
2 & \checkmark & \checkmark & & 16.51 & 35.05 & 18.01 & 36.02 & 20.40 & 42.99 & 18.31 & 38.02 & 28.16 \\
3 & & & \checkmark & 19.78 & 44.22 & 18.50 & 40.82 & 24.38 & 48.80 & 20.89 & 44.62 & 32.75 \\
4 & \checkmark & & \checkmark & 24.69 & 49.53 & 30.86 & 53.29 & 33.30 & 60.17 & 29.62 & 54.33 & 41.97 \\
5 & \checkmark & \checkmark & \checkmark & \textbf{28.16} & \textbf{53.45} & \textbf{33.81} & \textbf{57.80} & \textbf{37.28} & \textbf{63.44} & \textbf{33.08} & \textbf{58.23} & \textbf{45.65} \\
\bottomrule
\end{tabular}
\end{table*}

\subsection{Implementation Details}
For the base model, we use ResNet-50 \cite{He2016resnet} as the image encoder, and 12-layers BERT \cite{Devlin2019BERT} as the text encoder. We initialize them with the CLIP(RN50) \cite{Radford2021CLIP}.
For the large model, we upgrade the image encoder with Vision-Transformer \cite{Do2021ViT} and initialize the model with the CLIP(ViT-B/32).
The hidden dimensions of the semantic joint embedding space for our base and large models are set as 1024 and 512 respectively.
The learnable temperature $\tau_1$ and $\tau_2$ in the InfoNCE losses are both initialized as 0.07.
The hyper-parameter $\tau_3$ for pseudo label construction is set as 4.
The $\lambda$ in Eq.(\ref{eq:total_loss}) is set as 0.5.
We set the initial learning rate for pre-trained encoders as 1e-6, for the modules trained from scratch as 1e-4 in all the training stages.
As the training steps increase, we decay the learning rates.
The mini-batch size $B$ is set as 32 and we train the model with 100K and 20K iterations for stage-two and stage-three respectively.
We implement the whole model with PyTorch and conduct all the experiments on a single NVIDIA GeForce RTX 2080 Ti GPU.

\subsection{Comparison with State-of-the-art Methods}
We compare our model with the following state-of-the-art methods.
\begin{itemize}[leftmargin=18pt]
\setlength{\itemsep}{0pt}
\setlength{\parsep}{0pt} 
\setlength{\parskip}{0pt}
\item \textbf{TIRG} \cite{vo2019tirg} is the first model proposed for the CTI-IR task, which composes image and text query features through gating and residual connection to modify partial image features while keep others unchanged.
\item \textbf{Film} \cite{Perez2018Film} is proposed to inject text features into the image map from CNN by feature-wise affine transformation.
\item \textbf{Relationship} \cite{Santoro2017Relationship} concatenates the text and image feature maps and learns cross-modal relationship with MLP.
\item \textbf{VAL} \cite{chen2020val} composes image and text features at multiple layers of CNN by attention mechanism to capture the multi-scale image information and match them hierarchically. An auxiliary visual-semantic matching objective between the image and textual side information is also used to help learn the semantic embedding space better.
\item \textbf{DATIR} \cite{gu2021DATIR} is proposed to combine attention mutual information maximization and hierarchical mutual information maximization to bridge the modality gap.
\item \textbf{JPM} \cite{yang2021JPM} is proposed with an auxiliary module to align the difference between reference and target images with the modification text by joint prediction.
\item \textbf{CosMo} \cite{Lee2021CosMo} employs two modules including content modulator and style modulator to perform local and global updates to the reference image feature based on text.
\item \textbf{DC-Net} \cite{Kim2021DCNet} proposes a dual network including a composition module which is a variant of TIRG \cite{vo2019tirg}, and a correction module which computes the similarity between the difference representation of reference and target images and the modification text embedding.
\item \textbf{CLVC-Net} \cite{wen2021clvcnet} combines local-wise and global-wise composition modules for both local and global modification demands, and designs a mutual enhancement module to share the knowledge between each composition module.
\end{itemize}

Table~\ref{tab:fashioniq} reports the CTI-IR results from different methods on the Fashion-IQ dataset.
For a fair comparison with previous works, we report the results of our base model and large model on both retrieval candidate splits.
The VAL's candidate split contains fewer candidate images as shown in Table~\ref{tab:data_split}, and thus leads to higher retrieval performance than the original candidate split.
Our model significantly outperforms the state-of-the-art methods on both two splits, improving the Rmean by 22.9\% and 14.7\% respectively, even using only our base model.
It demonstrates the benefits of open-domain and fashion-domain single-query retrieval knowledge for the CTI-IR task, and the effectiveness of our multi-stage progressive learning strategy.
In addition, our proposed adaptive weighting strategy, which dynamically assigns importance weights on image and text queries in different retrieval cases, also brings improvements.
Our large model achieves additional gains compared with our base model, which shows the potential of our proposed method.
Similar results have also been achieved on the Shoes dataset as shown in the Table~\ref{tab:shoes}.
Both our base and large models outperform other state-of-the-art methods.

\subsection{Ablation Studies}
\begin{table*}[t]
\centering
\caption{Ablation study with different pre-training tasks and datasets in stage-two fashion-domain pre-training. Zero-shot setting denotes the model has not been adapted to the CTI-IR task by stage-three training. The results are averaged from the three (Dress, Shirt, Tops\&Tees) subsets.}
\label{tab:abalation_data}
\begin{tabular}{c|ccc|cc|ccc|ccc}
\toprule
\multirow{2}{*}{Row} & \multicolumn{3}{c}{Pre-training Data} & \multicolumn{2}{|c|}{Tasks} & \multicolumn{3}{c|}{Zero-shot} & \multicolumn{3}{c}{Fine-tuned} \\
& FashionGen & FACAD & DeepFashion & FIR & FAP & R@10 & R@50 & Rmean & R@10 & R@50 & Rmean\\
\midrule
1 & - & - & - & - & - & 9.57 & 18.98 & 14.27 & 29.62 & 54.33  & 41.97 \\
\midrule
2 & \checkmark & & & \checkmark & & 12.35 & 27.30 & 19.82 & 31.94 & 56.84 & 44.39 \\
3 & & \checkmark & & \checkmark & & 16.34 & 35.76 & 26.05 & 32.76 & 57.53 & 45.15 \\
4 & \checkmark & \checkmark & & \checkmark & & 17.35 & 36.92 & 27.14 & 32.67 & 58.09 & 45.38 \\
5 & & & \checkmark & & \checkmark & 6.42 & 14.00 & 10.21 & 32.87 & 57.82 & 45.35 \\
6 & \checkmark & \checkmark & \checkmark & \checkmark & \checkmark & \textbf{18.31} & \textbf{38.02} & \textbf{28.16} & \textbf{33.08} & \textbf{58.23} & \textbf{45.65} \\
\bottomrule
\end{tabular}
\end{table*}

\begin{table*}[t]
\centering
\caption{Ablation study with different composing methods for image and text queries on the Fashion-IQ dataset.}
\label{tab:abalation_composing}
\begin{tabular}{c|l|cccccc|ccc}
\toprule
\multirow{2}{*}{Row} & \multirow{2}{*}{Method} & \multicolumn{2}{c}{Dress} & \multicolumn{2}{c}{Shirt} & \multicolumn{2}{c|}{Tops\&Tees} & \multicolumn{3}{c}{Avg} \\
&& R@10 & R@50 & R@10 & R@50 & R@10 & R@50 & R@10 & R@50 & Rmean \\
\midrule
1 &Image-Only & 4.12 & 12.79 & 8.29 & 18.60 & 7.19 & 16.06 & 6.53 & 15.82 & 11.17 \\
2 &Text-Only & 19.93 & 41.84 & 25.66 & 45.63 & 28.76 & 53.49 & 24.78 & 46.99 & 35.89 \\
3 &Mean Pooling & 28.16 & 53.45 & 33.81 & 57.80 & 37.28 & 63.44 & 33.08 & 58.23 & 45.65 \\
4 &Concatenation & 24.14 & 50.72 & 24.09 & 46.17 & 28.81 & 56.96 & 25.68 & 51.28 & 38.48 \\
5 &Residual Gating & 22.91 & 48.04 & 22.52 & 45.00 & 29.02 & 55.02 & 24.81 & 49.35 & 37.08 \\
6 &Adaptive & \textbf{29.00} & \textbf{53.94} & \textbf{35.43} & \textbf{58.88} & \textbf{39.16} & \textbf{64.56} & \textbf{34.53} & \textbf{59.13} & \textbf{46.83} \\
\bottomrule
\end{tabular}
\end{table*}

In order to demonstrate the contributions from different components in our proposed model, we conduct several ablation studies on the Fashion-IQ dataset based on our base model.

\subsubsection{\textbf{Multi-stage Progressive Learning}}
Table~\ref{tab:abalation_progress} shows the effectiveness of different training stages in our mutli-stage progressive learning. 
In order to exclude the impact of the composing module, we only use mean pooling to compose image and text query features.
The Row~1 reports the results with only open-domain pre-training, which is the zero-shot performance of CLIP on the CTI-IR task.
Since CLIP is only pre-trained with the open-domain single-query image retrieval task, it achieves unsatisfactory results on the CTI-IR task with an average recall of 14.27. 
It further verifies the difficulty of the CTI-IR task which is quite different from the conventional image retrieval.
The Row~2 shows retrieval performance with two-stage pre-training including open-domain and fashion-domain pre-training with our proposed fashion-related tasks, which is better than the Row~1, although the model has also never been optimized by the CTI-IR task.
Surprisingly, it even outperforms the results of many previous works \cite{vo2019tirg, Perez2018Film} trained on the CTI-IR task and data, which demonstrates that the single-query domain-specific image retrieval knowledge is indeed helpful for the CTI-IR task, while it has not been well exploited in previous works.
With the full multi-stage learning, our model achieves the best results as shown in the Row~5.
Comparing the Row~3-5, it shows that single-query image retrieval pre-training in both open-domain and fashion-domain brings improvements to our model, and each training stage is necessary.

\subsubsection{\textbf{Pre-training Data and Tasks}}
Table~\ref{tab:abalation_data} ablates the impact of different pre-training data and tasks used for the stage-two fashion-domain pre-training on the final retrieval performance.
The zero-shot setting means that the model has not been adapted to the CTI-IR task by stage-three learning, under which the fashion-domain pre-training is most critical for the performance. The fine-tuned setting means that the model has additionally undergone the stage-three learning on the triplet-format data. 
The first row stands for the baseline without fashion-domain pre-training.
The Row 2-4 show the results using only the FIR task for pre-training.
Using only FashionGen or FACAD dataset, our model outperforms the baseline model under both zero-shot and fine-tuned settings.
When merging both datasets for the FIR task, the performance is further improved to the Rmean of 45.38.
The Row 5 shows the results using only the FAP task for pre-training on the DeepFashion dataset.
Although the FAP task is very different from the CTI-IR task, it also brings improvements with the Rmean of 45.35, which is competitive with the model pre-trained with the FIR task.
It demonstrates the effectiveness of our proposed FAP task.
When combining FIR and FAP pre-training tasks in the stage-two learning, our model achieves the best results with the Rmean of 45.65.

\subsubsection{\textbf{Query Composing Methods}}
In Table~\ref{tab:abalation_composing}, we ablate the model with different composing approaches for the hybrid-modality queries.
All of the models are equipped with the three-stage progressive learning strategy, only the composing method $\mathcal{F}(\cdot)$ is different.
The Row~1 and Row~2 are the image retrieval results using only the image or the text query in the hybrid-modality query, which are the models used for our pseudo importance weights generation as described in Section~\ref{sec:con2}.
The text-only model is shown to outperform the image-only model, which indicates that in most of retrieval cases, the text query is more important than the image query on the Fashion-IQ dataset.
When simply adopting the mean pooling as the composing method for the hybrid-modality queries, the model achieves improvement compared with the single query models of Row~1\&2, which indicates that multi-modal queries are complementary and it is necessary to compose them for better retrieval.
The Row~4 and Row~5 are the two common composing methods, namely concatenation and residual gating. 
The concatenation approach concatenates the image and text features and feed them into a MLP to get the fused query vector, which is commonly used for the multi-modal fusion in other cross-modal tasks.
However, it achieves much inferior results than the mean pooling.
Similarly, the residual gating approach which was used in TIRG \cite{vo2019tirg} also achieves worse retrieval results.
We consider the reason is that the complex fusion layers employed in previous methods will destroy the semantic joint embedding space learned through the first two stages, so that the learned knowledge cannot be fully exploited.
Our proposed self-supervised adaptive composing module however makes full use of the knowledge learned in the open-domain and fashion-domain pre-training stages and adaptively focuses on the two modality queries for different retrieval intentions, thus achieves the best results.



\begin{figure*}
  \centering
  \includegraphics[width=\linewidth]{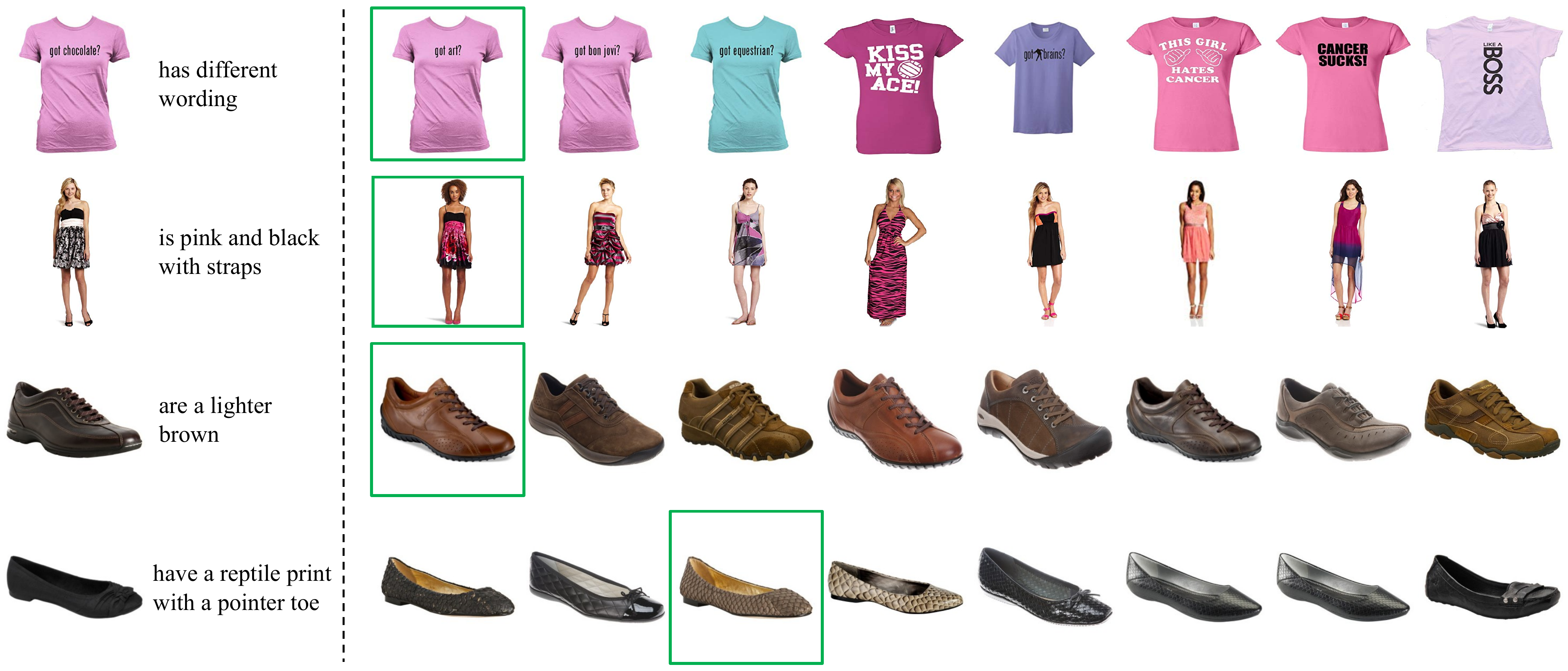}
  \caption{Qualitative results of our base model on the CTI-IR task. The first two rows are the examples from the Fashion-IQ dataset, and the last two rows are from the Shoes dataset. The ground-truth image is highlighted with green box.}
  \label{fig:retrieval_case}
\end{figure*}

\begin{figure}
  \centering
  \includegraphics[width=\linewidth]{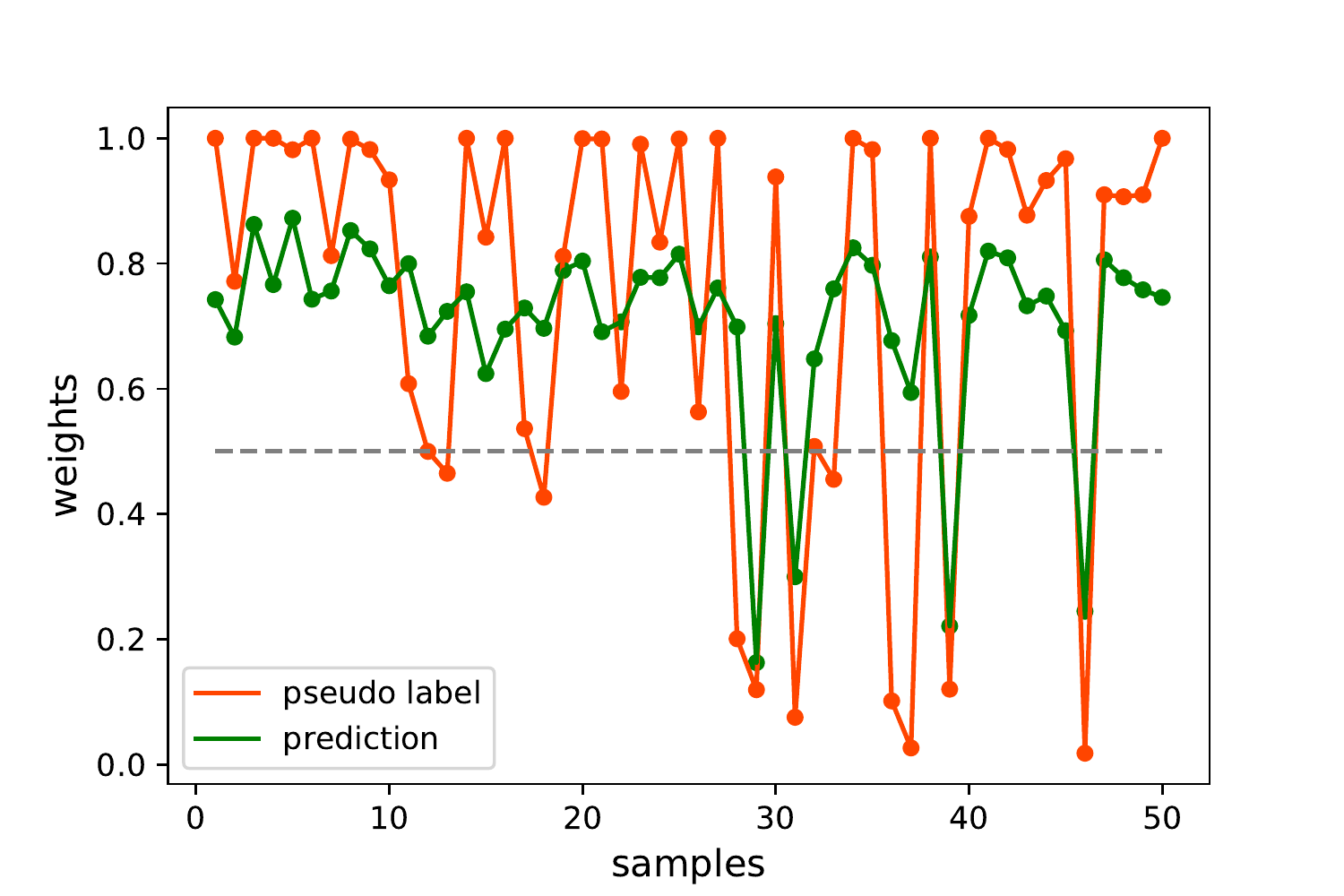}
  \caption{Visualization of the predicted weights and pseudo labels for text query on 50 instances from Fashion-IQ dataset. The weight for image query is 1 minus that of text query.}
  \label{fig:weights}
\end{figure}

\subsection{Qualitative Analysis}
In addition to the quantitative results, we also show some qualitative results to analyze the effectiveness of our proposed model.

\subsubsection{\textbf{Image Retrieval Results}}
In Figure~\ref{fig:retrieval_case}, we show some retrieval examples of our base model on the Fashion-IQ and Shoes datasets.
Our model can successfully retrieve the target image based on the combination of image and text queries.
For the first row, the model is asked to search an image very similar to the reference image except with different words on the shirt.
Our model successfully retrieves the target image and ranks it first.
Furthermore, we find that the image ranked in the second, although not the ground-truth target image, does satisfy the retrieval intention and is also ranked high by our model.
It demonstrates that our model has learned an effective semantic embedding space for image retrieval.
For the second row, a more detailed and descriptive modification demand is provided, which requires the model to fully consider both the image and text queries in order to correctly retrieve the target image.
Our model also performs well on such cases and returns multiple pink and black dresses with straps (the first, second and fourth ranks).
It demonstrates that our model can handle different retrieval demands by dynamically paying more attention on the image or text query.
The same results are also observed on the Shoes dataset.
In the last row, we show an example where the model does not retrieve the ground-truth image in the first rank.
Although the ground-truth target image is ranked in the third place, we find that the retrieved images in the top-4 all satisfy the retrieval demand, which further demonstrates the effectiveness of our model.

\subsubsection{\textbf{Adaptive Query Weighting}}
In Figure~\ref{fig:weights}, we visualize the predicted text query weights and pseudo labels in our self-supervised adaptive query weighting module.
We sample 50 instances from the Fashion-IQ dataset and show the weights of the text queries.
It is shown that for most of the retrieval instances in Fashion-IQ dataset, the text query plays a more important role in the hybrid-modality query image retrieval, which is consistent with the results that \emph{Text-Only} model performs better than the \emph{Image-Only} model in the Table~\ref{tab:abalation_composing}.
The predicted weights have a similar trend with the pseudo labels, which demonstrates that our model can appropriately adapt the modality importance weight for different retrieval intentions.
\section{Conclusion}
In this work, we focus on the more challenging image retrieval task that involves composing text and image queries for image retrieval (CTI-IR).
Due to the complexity of the task and the data scarcity of the <reference-image, modification-text, target-image> triplets, we propose to decompose this task into a three-stage learning problem to learn the knowledge step by step progressively.
Specifically, we first leverage the semantic joint embedding space for open-domain image retrieval task from the CLIP, and then transfer the knowledge to the fashion-domain with two proposed fashion-related pre-training tasks.
Finally, we adapt the single-query image retrieval knowledge to the hybrid-modality query scenario with self-supervised query adaptive composing module.
We construct pseudo weight labels for image and text queries to indicate which modality is more important in each retrieval scenario, and teach the model to dynamically focus on the two modality queries for different retrieval intentions.
Extensive experiments on two benchmark datasets Fashion-IQ and Shoes show that our model significantly outperforms state-of-the-art methods by 24.9\% and 9.5\% on the mean of Recall@K respectively.
Qualitative results also show that the importance of image and text queries are indeed different in different retrieval instances, while our model learns to dynamically assign importance weights for better target image retrieval.

\begin{acks}
This work was partially supported by National Natural Science Foundation of China (No.62072462) and National Key R\&D Program of China (No.2020AAA0108600).
\end{acks}

\bibliographystyle{ACM-Reference-Format}
\bibliography{reference}

\end{document}